\pdfoutput=1

\documentclass[runningheads]{llncs}
\usepackage{graphicx}
\usepackage{svg}
\usepackage{subcaption}
\usepackage{multirow}
\usepackage{diagbox}
\usepackage{booktabs}
\usepackage{adjustbox}
\usepackage{pdfpages}
\graphicspath{{figures/}}
\newcolumntype{C}{>{\centering\arraybackslash}m{0.4\textwidth}}

\begin{document}
\title{Selective Manipulation of Disentangled Representations for Privacy-Aware Facial Image Processing}
\titlerunning{Privacy-Aware Facial Image Processing}
%
\author{Sander De Coninck\inst{1}\and
Wei-Cheng Wang\inst{1} \and
Sam Leroux\inst{1} \and
Pieter Simoens\inst{1}}
\authorrunning{De Coninck et al.}
%
\institute{IDLab, Department of Information Technology at Ghent University - imec \\Technologiepark 126, B-9052 Zwijnaarde, Belgium \\
\email{{first.lastname}@ugent.be}\\
}
%
\maketitle              
\begin{abstract}
Camera sensors are increasingly being combined with machine learning to perform various tasks such as intelligent surveillance. Due to its computational complexity, most of these machine learning algorithms are offloaded to the cloud for processing. However, users are increasingly concerned about privacy issues such as function creep and malicious usage by third-party cloud providers. To alleviate this, we propose an edge-based filtering stage that removes privacy-sensitive attributes before the sensor data are transmitted to the cloud. We use state-of-the-art image manipulation techniques that leverage disentangled representations to achieve privacy filtering. We define opt-in and opt-out filter operations and evaluate their effectiveness for filtering private attributes from face images. Additionally, we examine the effect of naturally occurring correlations and residual information on filtering. We find the results promising and believe this elicits further research on how image manipulation can be used for privacy preservation.

\end{abstract}
\section{Introduction}
Recently, there has been an increase in popularity in using camera sensors and cloud services to capture and process large amounts of data. The benefits are clear, cameras are cheap, and cloud computing allows for easy maintainability. However, this comes at the cost of additional security and privacy risks. As the footage these cameras capture is rich in information, there is a considerable risk of function creep~\cite{marxMissionCreepSmart2020,WhatWrongPublic2002}(i.e., using the data for other purposes than originally intended). A camera used to monitor street activity can be misused for racial profiling as the footage captured is rich enough to do both. Another key concern is possible malicious use by the third-party cloud provider. To alleviate this, we propose an edge-based filtering stage that removes privacy-sensitive attributes before the sensor data is transmitted to the cloud. As such, the cloud server should only have access to the strictly required properties for the utility task. For optimal usability, we identify three key requirements. Firstly, it should provide a clear view of what features of the data are removed or kept. In other words, the filtering should be semantically interpretable. Additionally, there should be some flexibility as to what data is allowed or prohibited. This flexibility would allow the camera footage to be used for multiple purposes, where each task only gets the strictly necessary features. Lastly, a solution where no modifications of the data processing systems are required is appealing as well, as this allows the use of pre-existing and well-maintained API's and resources for the utility task.

\begin{figure}
    \centering
    \includegraphics[width=\linewidth]{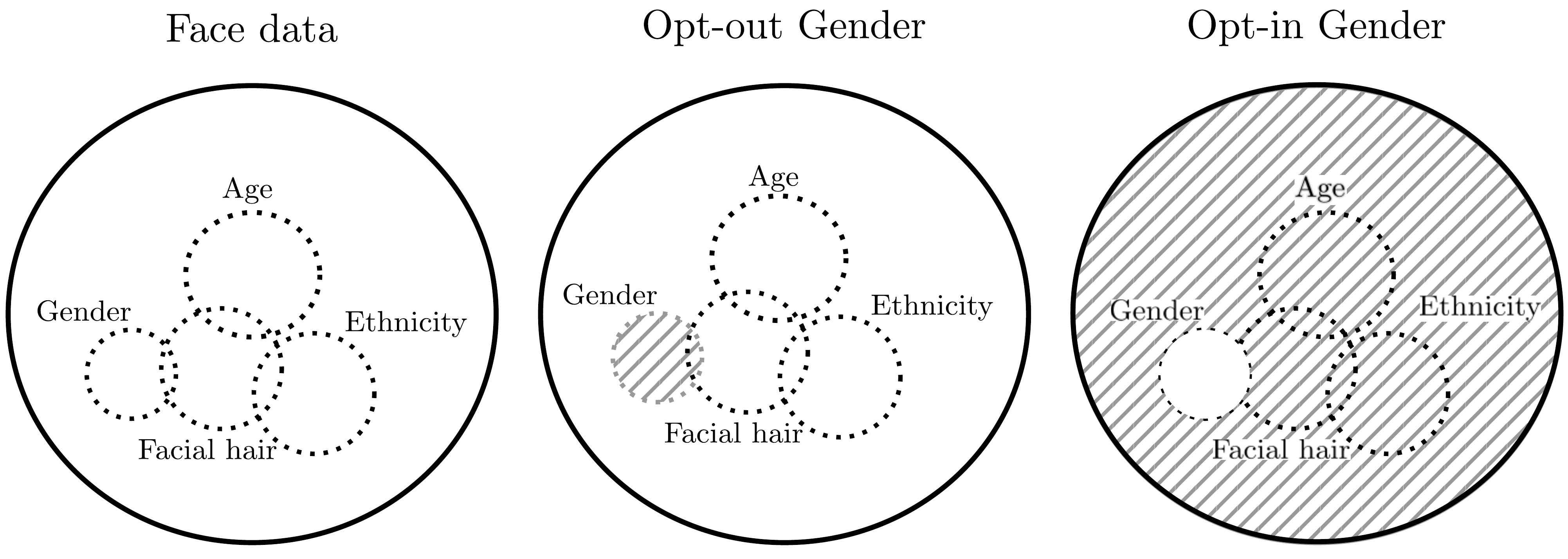}
    \caption{Illustration of opt-in and opt-out filtering for the gender attribute for face data. A striped pattern indicates the removal or change of data.}
    \label{fig:optin-optout}
\end{figure}

We propose to use the state-of-the-art image manipulation technique ZeroDIM~\cite{gabbayImageWorthMore2021a}, which is based on disentangled representation learning, to implement privacy filters. ZeroDIM can make high-quality edits of images by disentangling them into certain semantic factors. As such it should allow the necessary flexibility and interpretability of privacy protection. We can use this model to either explicitly state what data should be removed or alternatively state what data should be kept. Explicitly stating what features should be hidden is commonly referred to as opt-out. The more secure alternative, where the allowed features are explicitly stated, is called opt-in. A simple clarification of these operations can be seen in Figure \ref{fig:optin-optout}. Here we show how an image of a human face captures a combination of underlying attributes, such as age, gender and ethnicity. Some of these attributes partially overlap (e.g., facial hair and gender). Our goal is to transform this image into another image that either hides one attribute (opt-out) or retains a single attribute while hiding all others (opt-in).  

In this paper, we experimentally verify the use of the ZeroDIM image manipulation algorithm to implement privacy filters. We focus on facial imagery since the processing of face images is a use case highly relevant to privacy-preserving machine learning due to the sensitive nature of human faces. Additionally, facial imagery is one of the domains in which image manipulation is most successful because of the relative ease of disentanglement. 

This paper is organized as follows: Section \ref{sec:rel_works} describes related works in image manipulation and privacy-preserving machine learning. In Section \ref{sec:zerodim} we describe the ZeroDIM model and propose how to use it as a means of filtering. We then evaluate opt-in and opt-out settings for this purpose in Section \ref{sec:filtering}. Section \ref{sec:investigations} investigates some of the inherent difficulties of filtering face attributes, such as correlations between attributes and residual information. Finally, we conclude our work and look at future research in Section \ref{sec:conclusion} and \ref{sec:future}.

\section{Related Works}
\label{sec:rel_works}

\subsection{Image manipulation}
A growing body of literature has investigated the powers of pre-trained unconditional generators such as StyleGAN~\cite{karrasStyleBasedGeneratorArchitecture2019,karrasAnalyzingImprovingImage2020a,karrasAliasFreeGenerativeAdversarial2021} for semantic image manipulation/editing. A partition of these methods leverages the naturally disentangled latent space of pre-trained GAN models to develop latent manipulations that allow for specific semantic operations~\cite{harkonenGANSpaceDiscoveringInterpretable2020,shenInterpretingLatentSpace2020a}. When coupled with a GAN inversion technique ~\cite{roichPivotalTuningLatentbased2021,abdalImage2StyleGANHowEmbed2019,alalufReStyleResidualBasedStyleGAN2021,tovDesigningEncoderStyleGAN2021}, which embeds images into the latent space of these GAN models, this allows for powerful image manipulation, especially in the domain of facial imagery.

While these methods allow high-quality image editing, finding the wanted edit requires manual human examination or a set of annotated data for each edit. Because of this, others turn to natural language for text-guided image manipulation~\cite{xiaTediGANTextGuidedDiverse2021}. Research includes StyleCLIP~\cite{patashnikStyleclipTextdrivenManipulation2021}, which uses CLIP~\cite{radfordLearningTransferableVisual2021a} for guiding the latent manipulation, as well as ZeroDIM~\cite{gabbayImageWorthMore2021a}, which uses disentangled representation learning and leverages CLIP as a form of weak supervision. 

Although the field of image manipulation is already well established, only a few works have investigated their use for privacy insurance of face images. Most notable is PrivacyNet~\cite{mirjaliliPrivacyNetSemiadversarialNetworks2020}, a model that tries to hide face attributes while retaining identity information. Wang et al~\cite{wangFacePrivacyProtection2021} has a similar goal. Both works explicitly feed the new values for the face attributes to the generator to create filtered images. As such, they are limited by knowing the attribute values before filtering. Cao et al.~\cite{caoPersonalizedInvertibleFace2021} achieve the opposite goal. They remove identity information while keeping soft biometrics such as age and gender. This paper focuses on the different face attributes and how these can be changed in relation to each other. Additionally, we employ disentangled representations for all attributes (Cao et al. also use disentanglement; however it is limited to disentangling identity from attributes). An important distinction is that we do not need ground truth labels to filter attributes and provide privacy by randomising instead of explicitly changing.

\subsection{Privacy-preserving machine learning}
Privacy-preserving machine learning is an area that has evoked a significant amount of interest over the years. In this section, we will highlight some of the most popular techniques that can be used for privacy-preserving machine learning applications. For a complete overview, we refer to~\cite{mireshghallahPrivacyDeepLearning2020}.

Homomorphic encryption (HE) is a technique that allows computation over encrypted data without requiring decryption by the data processor. This allows leveraging of the benefits of cloud-based computation while mitigating privacy risks as only the data owner holds the decryption key. Examples of HE include CryptoNets~\cite{gilad-bachrachCryptoNetsApplyingNeural2016} and GAZELLE~\cite{juvekarGazelleLowLatency2018}. The main downfall of HE is its heavy computational footprint. Although significant improvements have been made over the years, real-time processing remains infeasible~\cite{mireshghallahPrivacyDeepLearning2020}.

Secure multi-party computation (SMC) ensures privacy by splitting computations over several computing parties. Services can only be completed if all parties participate, thus ensuring privacy if one party is trusted. This technique suffers from the same problem as HE in that the additional computation and communication needed infers a heavy inference time cost. Some examples of using SMC for machine learning are MiniONN~\cite{liuObliviousNeuralNetwork2017}, Crypten~\cite{knottCrypTenSecureMultiParty2021}, and XONN~\cite{riaziXONNXNORbasedOblivious2019a}.

Differential privacy~\cite{dworkDifferentialPrivacySurvey2008} is arguably the most well-known technique for privacy insurance. Privacy is guaranteed by adding noise to individual data items to prevent attackers from being able to infer the presence of a particular item in a dataset. However, most work in differential privacy focuses on dataset anonymisation and tabular data instead of inference on high dimensional image data~\cite{mireshghallahPrivacyDeepLearning2020}.

Lastly, a group of information-theoretic techniques degrade all information not necessary for a specified non-sensitive task~\cite{mireshghallahPrivacyDeepLearning2020,malekzadehMobileSensorData2019,malekzadehPrivacyUtilityPreserving2020,malekzadehProtectingSensoryData2018,ravalOlympusSensorPrivacy2019}. The focus of most of these works is sensory data, however recently works for computer vision have been published~\cite{mireshghallahShredderLearningNoise2020a,osiaDeepPrivateFeatureExtraction2018} as well. Some earlier research focuses on providing an opt-in approach, however they lack semantic interpretability~\cite{lerouxPrivacyAwareOffloading2018a,deconinckPrivacyAwarePerson2021a} making it difficult to verify privacy claims. 

\section{Image manipulation for privacy}
\label{sec:zerodim}

\begin{figure}[h]
    \centering
    \includegraphics[width=\linewidth]{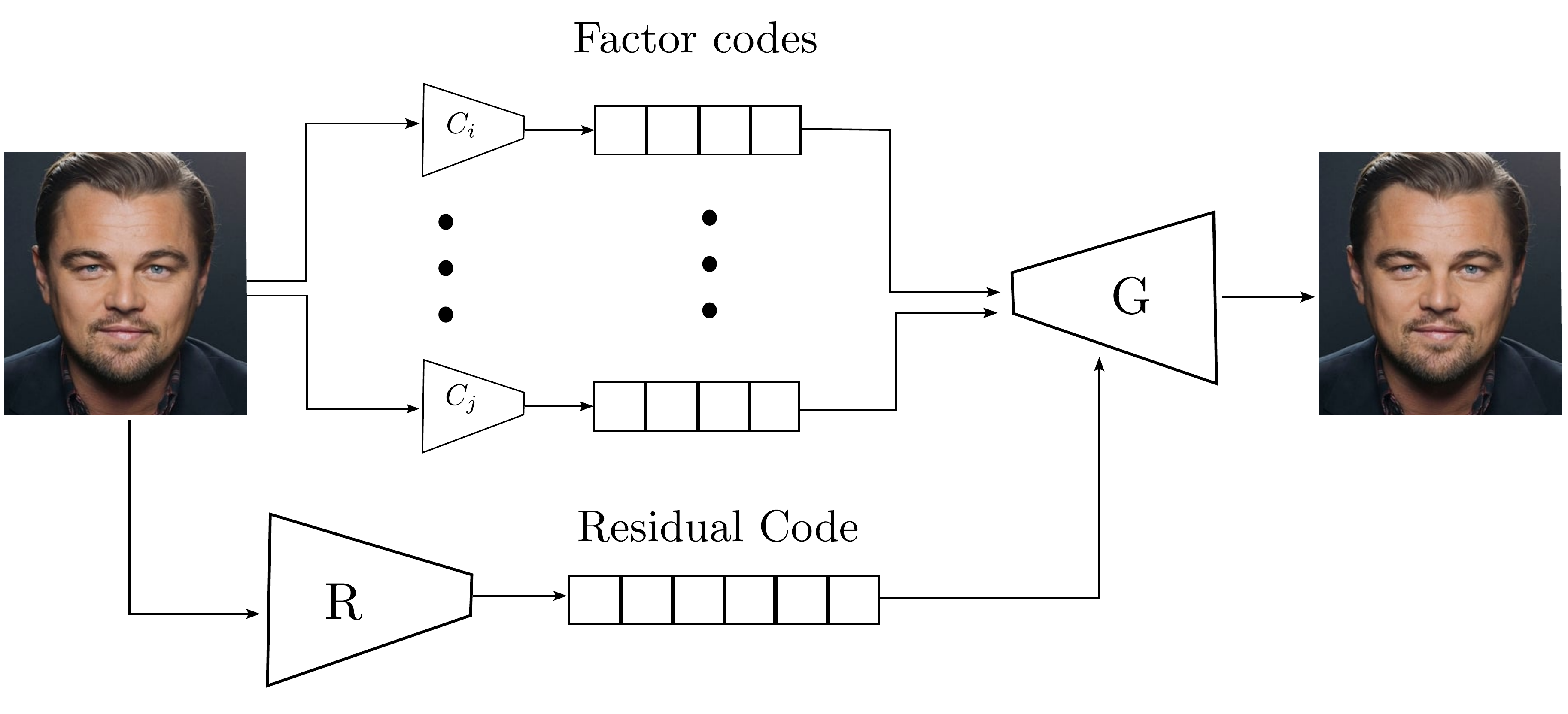}
    \caption{Architecture of ZeroDIM model, adapted from \cite{gabbayImageWorthMore2021a}. Disentangled factor codes are obtained through classifiers $C$, coupled with a residual code from the residual encoder $R$ and fed through the generator $G$ to reconstruct an image.}
    \label{fig:zerodim-arch}
\end{figure}

To investigate image manipulation as a means of filtering data, we utilize ZeroDIM (Zero-shot Disentangled Image Manipulation)~\cite{gabbayImageWorthMore2021a} as our manipulation model. We chose this model primarily due to its state-of-the-art results on facial data, a domain which lends itself well to privacy-preserving research. Additionally, due to its use of disentangled representations, the architecture forms an excellent basis for allowing both opt-in and opt-out settings.

ZeroDIM uses a weakly supervised disentanglement technique to create latent codes for a few predefined factors. The architecture of ZeroDIM can be seen in Figure \ref{fig:zerodim-arch}. It uses a set of classifiers $C$ for each attribute or factor and translates their predictions into distinct factor codes. Moreover, a residual encoder $R$ is used to obtain a residual code which contains all additional information (such as background or non-disentangled attributes) needed for accurate recreation. Both the factor codes and the residual code are used by a StyleGAN generator $G$ to reconstruct the image. The factor and residual codes are explicitly trained to have as little informational overlap as possible.

During training, distinct factor codes are learned for every factor value in the form of vector representations. At inference, the classifier of an attribute (e.g. gender), predicts a class (e.g. male). Subsequently, the factor code of that class is chosen and used in the generation process. This makes it so manipulating an attribute can be done by switching the factor code of that attribute with the factor code of another class. Due to the disentangled nature of these codes, switching a factor code should effectively remove all information of that attribute, making it a suitable operation for filtering.

\subsection{Filtering with ZeroDIM}

\begin{figure}[h]
    \centering
    \includegraphics[width=0.8\linewidth]{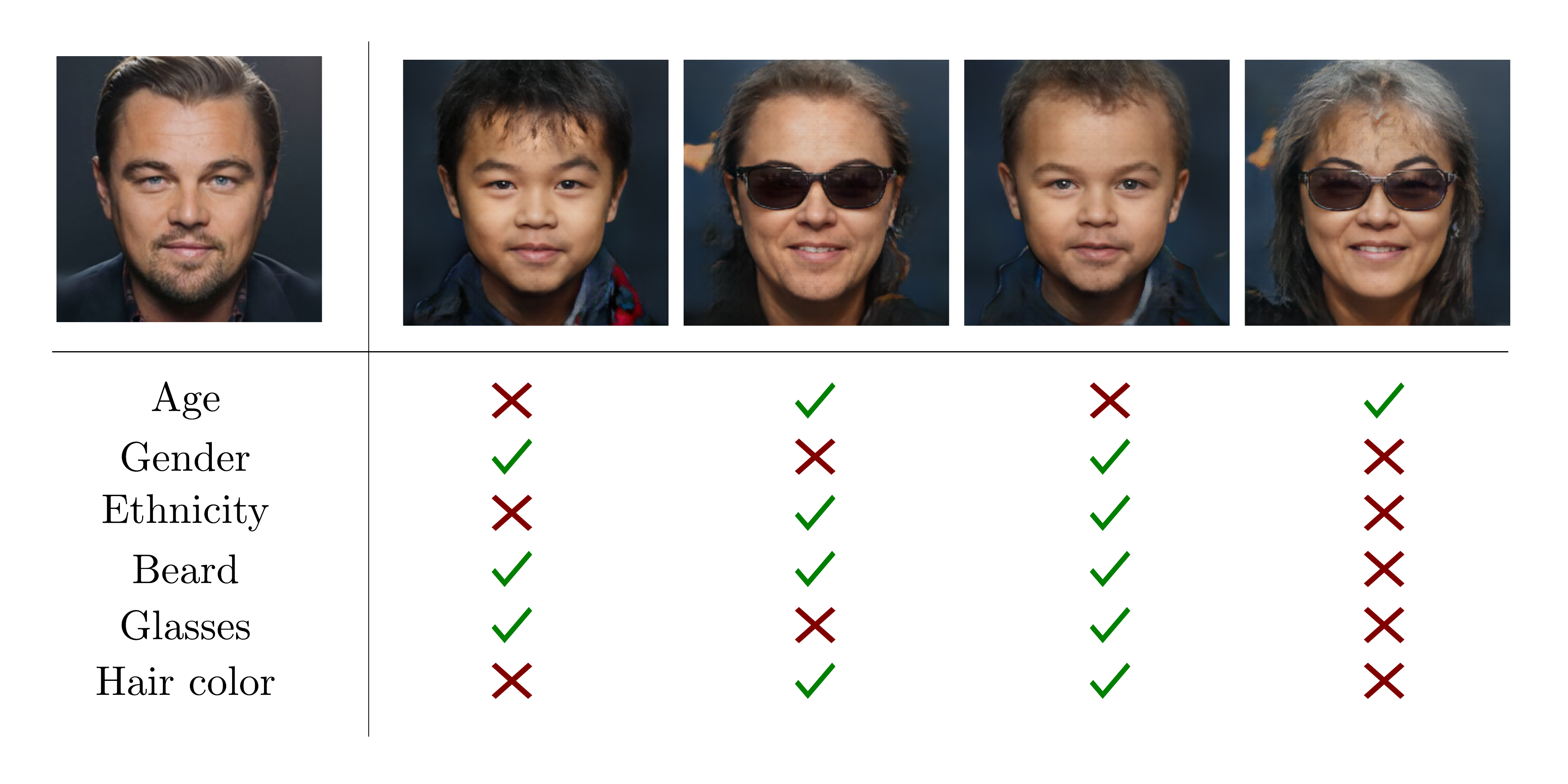}
    \caption{Example of filtering using the ZeroDIM model. The leftmost image shows the original version. Subsequent images show the results of filtering different combinations of allowed/disallowed attributes}
    \label{fig:ex-manip}
\end{figure}

We propose to make use of the disentangled nature of ZeroDIM representations to implement a privacy filter. At runtime, we replace the extracted feature vector of a privacy-sensitive attribute with a randomly selected factor code. For example, if we wish to filter gender, we replace the gender representation by choosing at random between those of male or female. Some example results can be seen in Figure \ref{fig:ex-manip}. We show the original face image on the left and subsequently the filtered images for different combinations of allowed attributes. For example, the first filtered image still allows detection of gender, beard and glasses, while hiding age, ethnicity and hair color. Note that the quality of the reconstructed images remains high even when multiple attributes are changed. 

\section{Experiments}
\label{sec:filtering}

\begin{figure}[h]
\centering
\begin{subfigure}{.5\textwidth}
  \centering  
  \includegraphics[scale=0.27]{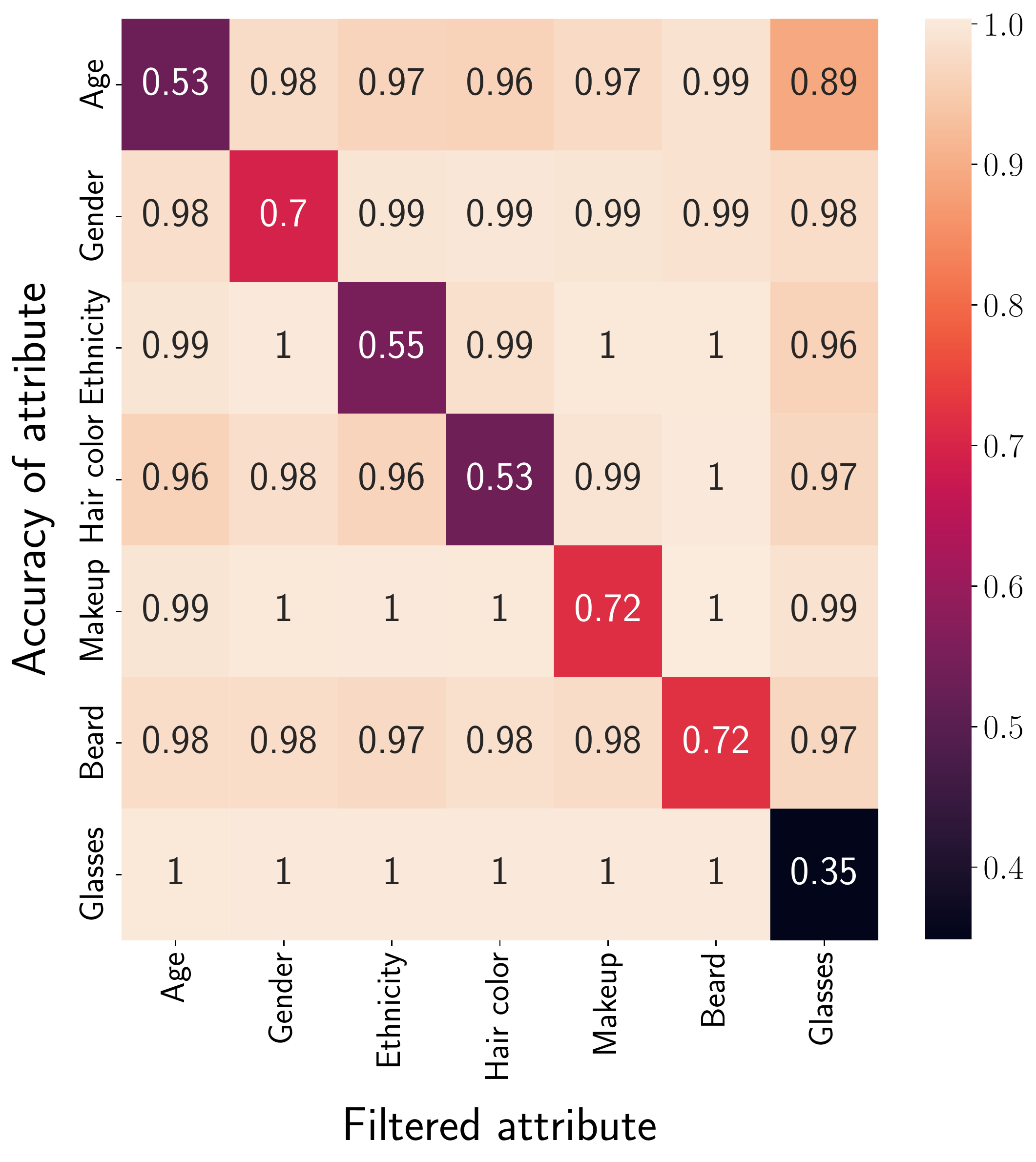}
    \caption{Opt-out}
    \label{fig:acc-oo}
\end{subfigure}%
\begin{subfigure}{.5\textwidth}
  \centering
  \includegraphics[scale=0.27]{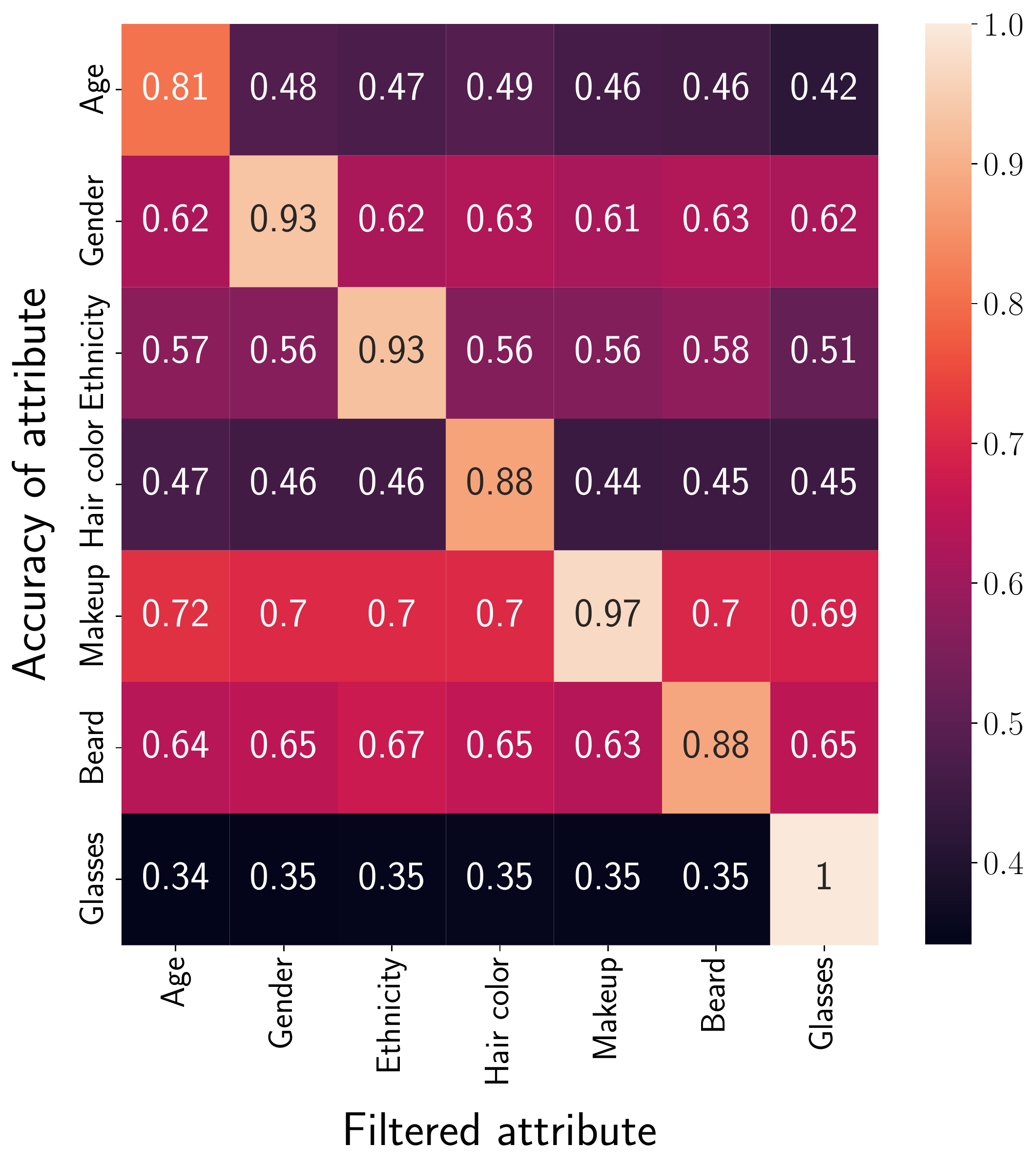}
    \caption{Opt-in}
    \label{fig:acc-oi}
\end{subfigure}
\caption{Relative accuracy of filtering when using an opt-out or opt-in configuration on a specified attribute. Columns indicate the attribute that was filtered, rows the attribute that was classified.}
\label{fig:acc_initial}
\end{figure}

In this section, we evaluate the effect of filtering on the classification accuracy of face attributes. Our method aims to transform a face image into a new face where some attributes have been replaced with random values. We use the pre-trained image classification models from ZeroDIM to obtain ground-truth values for all attributes. We then compare the accuracy of these models when recreating the faces with and without filtering. The optimal result is obtained when accuracy remains unchanged for an allowed attribute while degrading severely for disallowed attributes. For ease of interpretation, we limit ourselves to filtering single attributes. However, this is not a limitation of our approach. As shown in Figure \ref{fig:ex-manip}, we can use an arbitrary combination of allowed/disallowed attributes for filtering. For these experiments, we used the FFHQ~\cite{karrasStyleBasedGeneratorArchitecture2019} dataset, the same that was used for training ZeroDIM.

Figure \ref{fig:acc-oo} shows the result of the opt-out filtering. Here we aim to hide one attribute while still allowing accurate detection of the others. Each entry shows the relative accuracy of the column attribute when filtering the row attribute. Ideally, non-diagonal elements should have a relative accuracy of 1 (i.e., unchanged compared to unfiltered images), while elements of the diagonal should have significantly lower accuracy. The lower bound of these accuracies depends on the number of categories for each attribute. For example, gender has two categories (male and female), while age has four (child, teen, adult and elderly). Figure \ref{fig:acc-oo} confirms that our model is indeed able to hide the protected attribute while other information can still be accurately extracted. However, the accuracy of the filtered attribute is still not completely random, indicating that some information remains present in the residual code or other attributes. The only exception is the glasses attribute, we assume this to be because of the ease of disentanglement of this attribute as glasses are highly localized and easy to distinguish from other attributes. Another interesting observation is that filtering glasses causes a large drop in accuracy for the age attribute, while filtering any of the other factors does not. This is logical as glasses might be a good indicator of age. However, it raises the question of how related attributes influence filter operations. 

Figure \ref{fig:acc-oi} shows the results of the opt-in filter. Here we only allow one attribute to be extracted while removing all others. Ideally, the relative accuracy on the diagonal should be close to 1, while the others should be much lower. The results indicate that opt-in filtering results in lower accuracy for all attributes, those that were filtered out but also those that were not. Additionally, filtered accuracy still does not reach random values. We hypothesize that this could be due to two reasons. The first is that some information is still captured in the residual code, as such leaving this residual unchanged results in some data leakage. The second is that there is some information overlap in the different factors due to naturally occurring correlations. As a result, filtering out one correlated attribute but not the other allows some data leakage. We will extensively investigate these two causes in Section \ref{sec:investigations}.

\section{Investigating the causes of imperfect filtering}
\label{sec:investigations}
\subsection{Correlations between facial attributes}
\label{sec:corr}
In the previous sections, we assumed that ZeroDIM could disentangle the various face attributes perfectly. However, this is not realistic as there will always be some entanglement between attributes that are related in nature. For example, men are more likely to have facial hair than women and older people are more likely to have glasses than children. As stated in the last section, we assume this affects filtering as filtering out one attribute might still leave it deducible by using others. 

To identify the effect of these natural relations on filtering, we start by mapping the correlation between the different facial attributes. As these are categorical variables, we utilize Cramér's V~\cite{cramer2016mathematical} and the uncertainty coefficient~\cite{theilEstimationRelationshipsInvolving1970}. Both of these give insights into the correlation of discrete values.

\subsubsection{Cramér's V}

Cramér's V~\cite{cramer2016mathematical} is a metric for the correlation of categorical values based on Pearson's chi-squared statistic. It is symmetric and varies from 0 to 1. Cramér's V is defined as follows, with $\chi^2$ being Pearson's chi-squared test on the contingency table of $X$ and $Y$, $n$ the amount of samples, and $k$ and $r$ the number of categories of $X$ and $Y$ respectively:
\begin{equation}
    V(X,Y) = \sqrt{\frac{\frac{\chi^2}{n}}{min(k-1, r-1)}}
\end{equation}

Interpretation of the values is dependent on the number of degrees of freedom. A Cramer's V score of $0.15$ signifies a medium correlation when there are four degrees of freedom and only a small to medium correlation when fewer. A categorization can be found in Table \ref{tab:cramerv_cat}. 

\begin{table}[h]
    \centering
    \caption{Categorisation of Cramér's V correlations based on the degrees of freedom}
    \begin{tabular}{*4c}
    \toprule 
    Degrees of freedom & Small & Medium & Large\\ \midrule
     1 & 0.10 & 0.30 & 0.50 \\
     2 & 0.07 & 0.21 & 0.35 \\
     3 & 0.06 & 0.17 & 0.29 \\
     4 & 0.05 & 0.15 & 0.25 \\
     5 & 0.04 & 0.13 & 0.22 \\ \bottomrule
    \end{tabular}
    \label{tab:cramerv_cat}
\end{table}

\subsubsection{Uncertainty coefficient}
The second metric we use is the Uncertainty Coefficient or Theil's $U$~\cite{theilEstimationRelationshipsInvolving1970}, which is based on information theory. With $H$ as entropy and $I$ mutual information, we can define the uncertainty coefficient $U$ for $X$ given $Y$ as follows:
\begin{equation}
    U(X|Y) = \frac{H(X) - H(X|Y)}{H(X)} = \frac{I(X;Y)}{H(X)} 
\end{equation}The uncertainty coefficient is asymmetric and returns values between 0 and 1. Its information theoretical nature allows for intuitive interpretation. $U(X|Y)$ is the relative amount of information of $X$ we can infer from $Y$. Alternatively, knowing $Y$ results in a $U(X|Y)$ per cent reduction of uncertainty for $X$.

\begin{figure}[h]
\centering
\begin{subfigure}{.5\textwidth}
  \centering
  \includegraphics[scale=0.27]{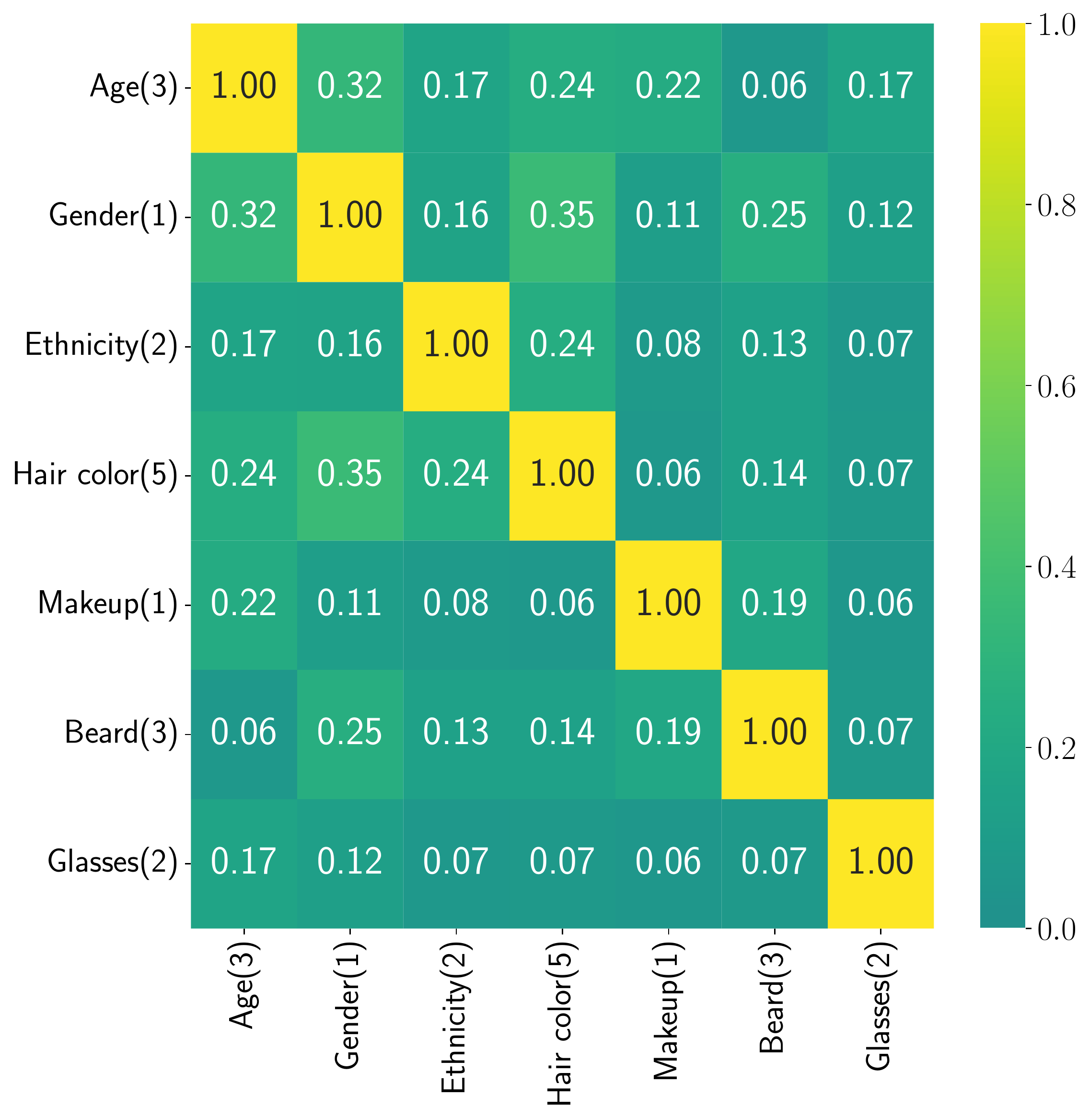}
    \caption{Cramér's V}
   \label{fig:cramer}
\end{subfigure}%
\begin{subfigure}{.5\textwidth}
  \centering
  \includegraphics[scale=0.27]{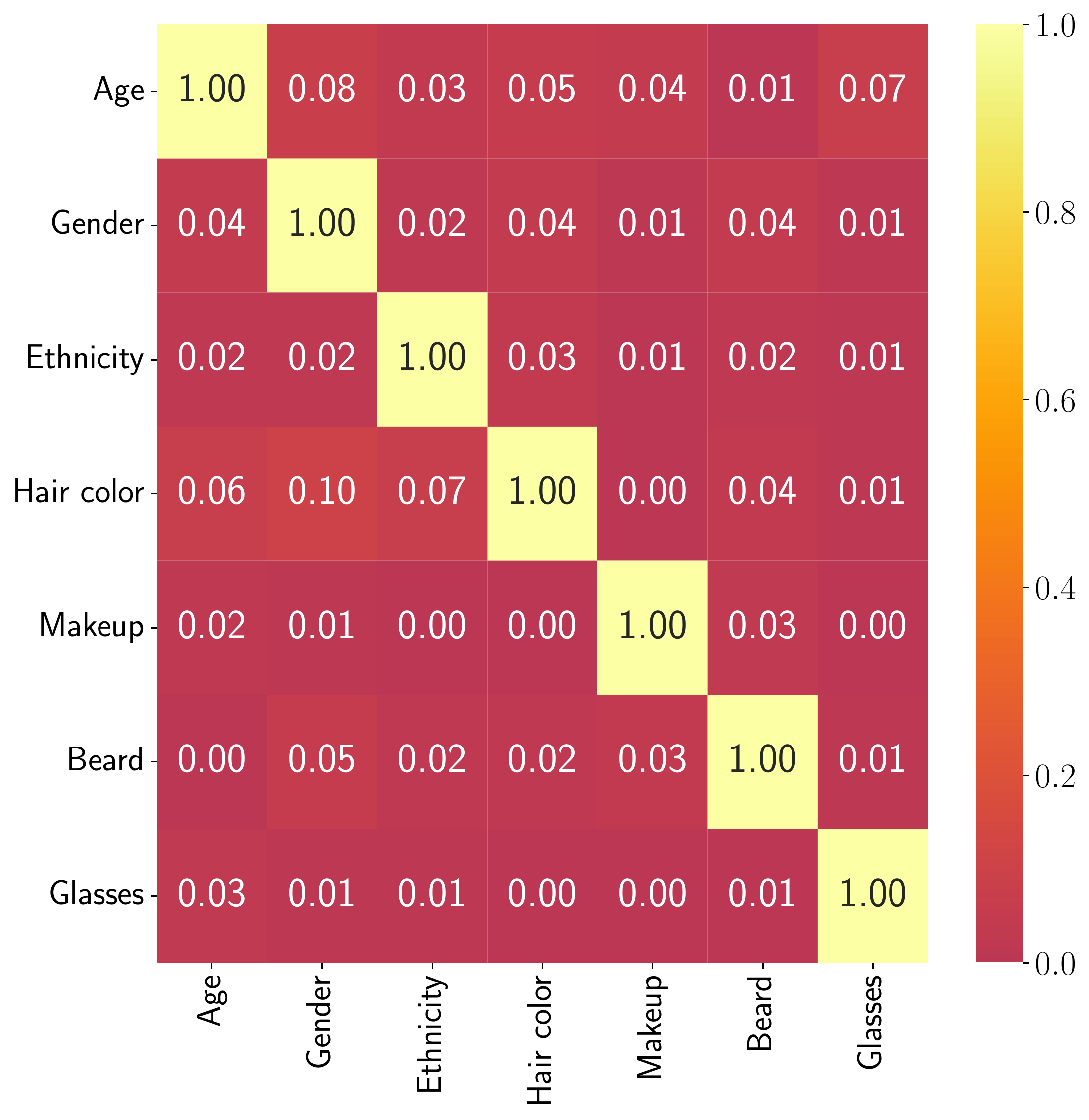}
    \caption{Uncertainty coefficient}
    \label{fig:uncertainty}
\end{subfigure}
\caption{Correlation matrices of the different facial factors using the Cramér's V and uncertainty coefficient metrics. Both have a range of 0 to 1 where higher values indicate more correlation between both factors. In the case of the uncertainty coefficient, one should read it as $U(row|column)$. For Cramér's V the number between parentheses indicates the degrees of freedom.}
\label{fig:correlations}
\end{figure}

Using these metrics, we calculated the correlations between all different facial attributes of images in the FFHQ dataset using the predictions of the pre-trained classifiers included in ZeroDIM. The results can be seen in Figure \ref{fig:correlations}. The figure on the left shows the result of Cramér's V score. For each different attribute, we show the degrees of freedom (number of classes - $1$ ). These can be used in conjunction with Table \ref{tab:cramerv_cat} to classify the strength of the correlation. Note that, for a given combination of two attributes, we should use the minimum of the degrees of freedom. For example, the minimum degrees of freedom of glasses and gender is $1$, while their correlation is $0.12$. Indicating that the relation between these two attributes is small according to Table \ref{tab:cramerv_cat}. Based on this categorisation, we can identify four medium strength relations of the non-diagonal elements in Figure \ref{fig:cramer}: age and gender, ethnicity and hair color, hair color and gender, and hair color and age. The first relation seems less intuitive than the latter three since gender and age have no direct correlation regarding appearance. However, this might be attributed to inherent biases or imbalances in the dataset. 

For the uncertainty coefficient results in Figure \ref{fig:uncertainty} similar conclusions arise, although the asymmetrical nature shows some interesting results. For example, knowing one's age gives less information about one wearing glasses as vice versa.

\subsection{The effects of correlation on filtering}
\label{sec:corr_and_filter}

To assess whether the correlation between attributes affects filtering quality we related the results of Section \ref{sec:filtering} to those of Section \ref{sec:corr}. This was done by testing the linear correlation between the accuracy difference (i.e., the accuracy of Y without filtering - the accuracy of Y when filtering X) and the correlation between X and Y. For an opt-out operation, we expect the relationship to be positive: when two attributes are related and we filter one, the other should drop in accuracy. The inverse hypothesis holds for opt-in. If we keep an attribute that is heavily related to the other, the drop in accuracy should be lower. 
\begin{table}
    \centering
    \caption{Regression plots indicating the relation between the drop in accuracy of Y when filtering X and the correlation of X and Y. A plot is shown for each combination of the correlation metric and filtering mode}
    \begin{tabular}{l C C}
       \toprule
         & Opt-out & Opt-in \\
        \midrule
        Cramér's V & \includegraphics[width=0.4\textwidth]{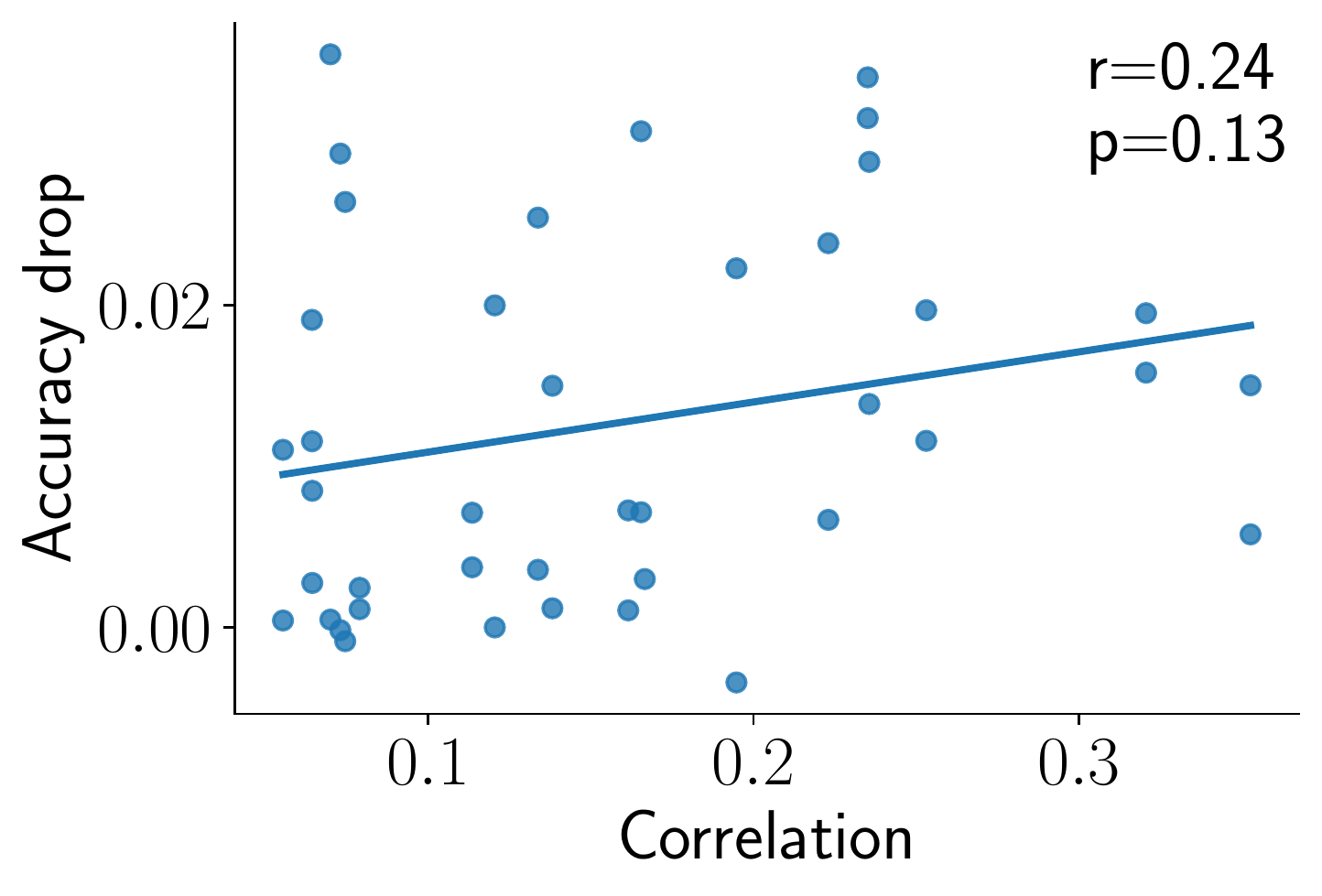} & \includegraphics[width=0.4\textwidth]{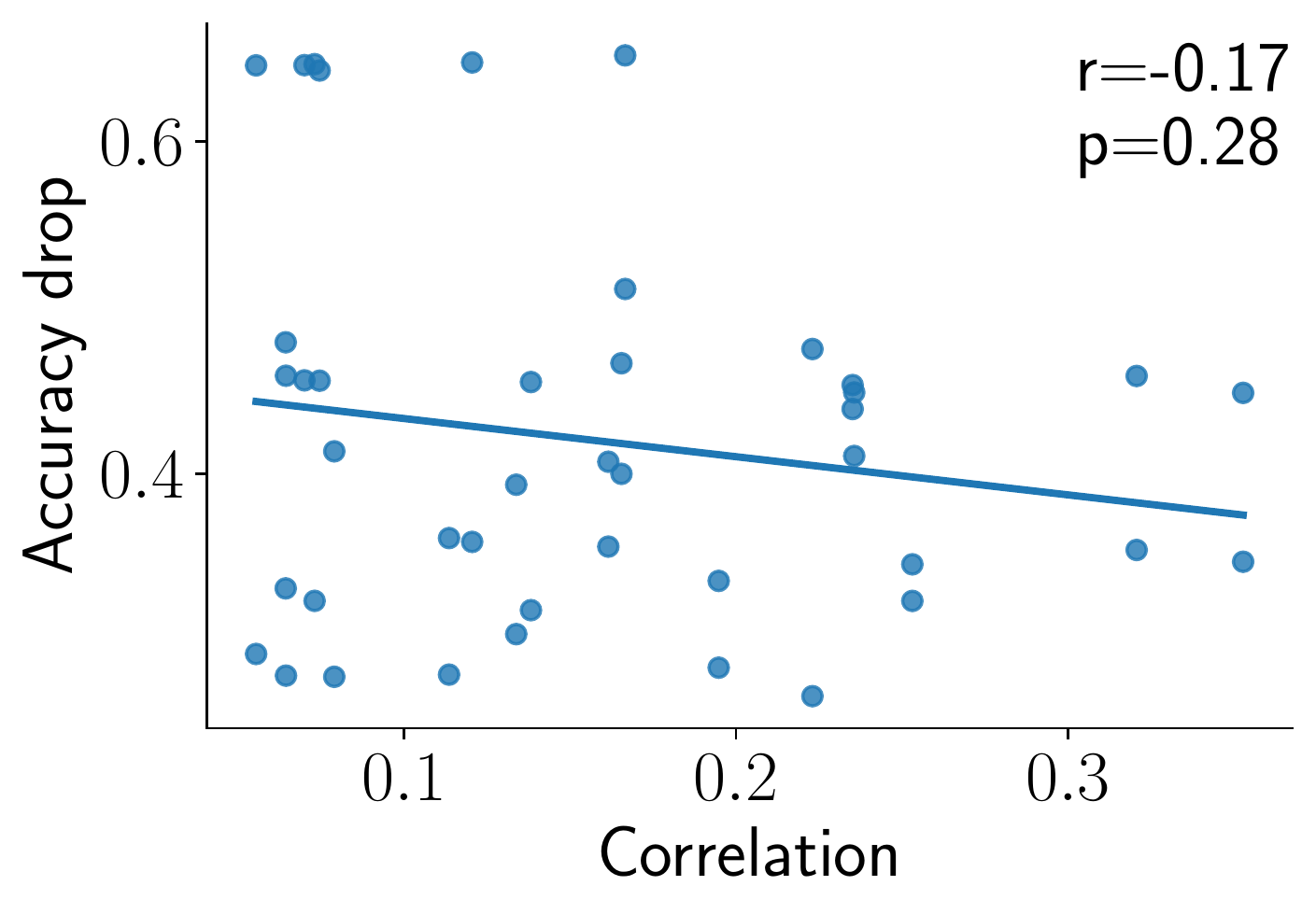} \\
        \shortstack{Uncertainty \\ coefficient} & \includegraphics[width=0.4\textwidth]{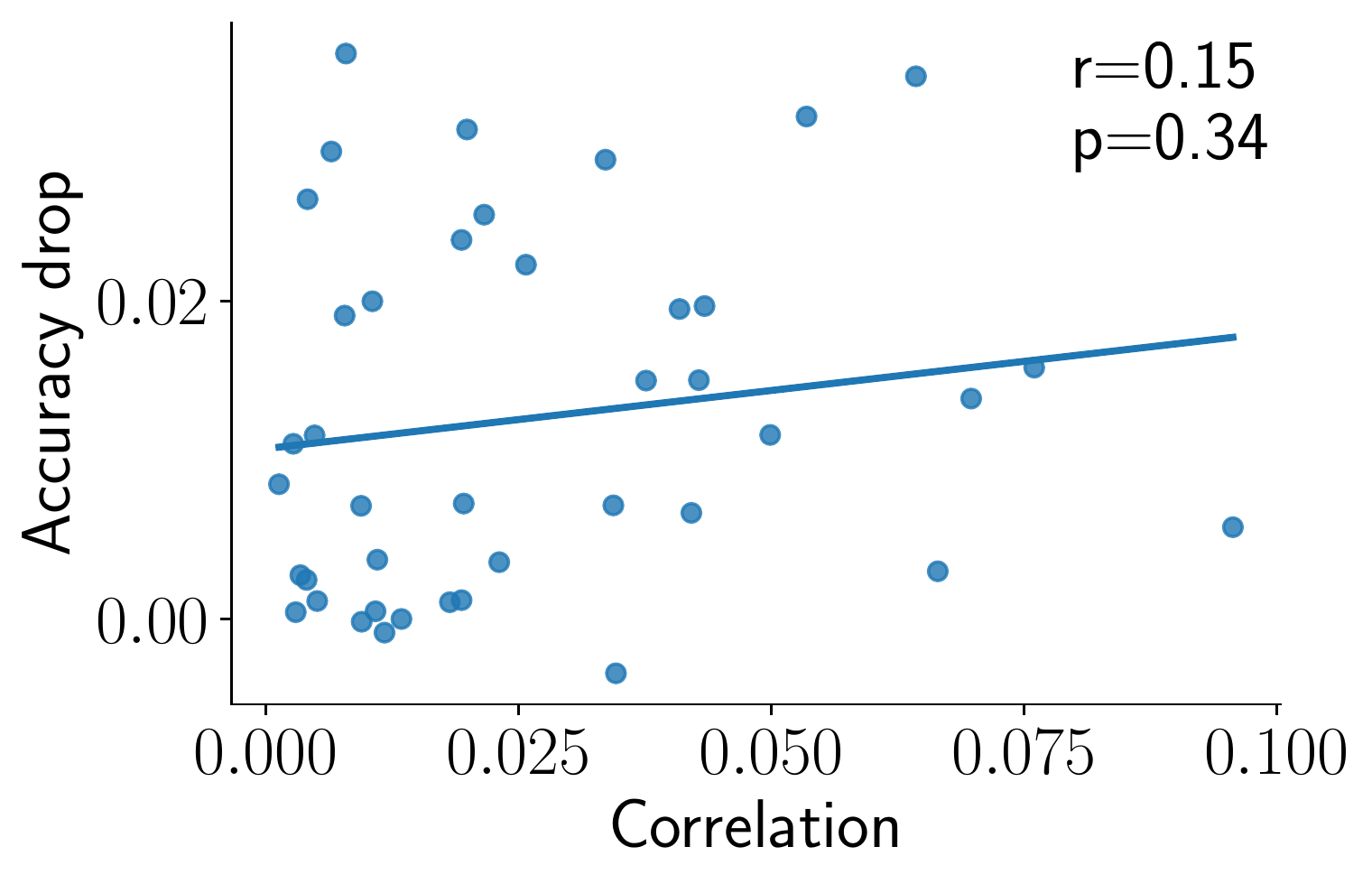} & \includegraphics[width=0.4\textwidth]{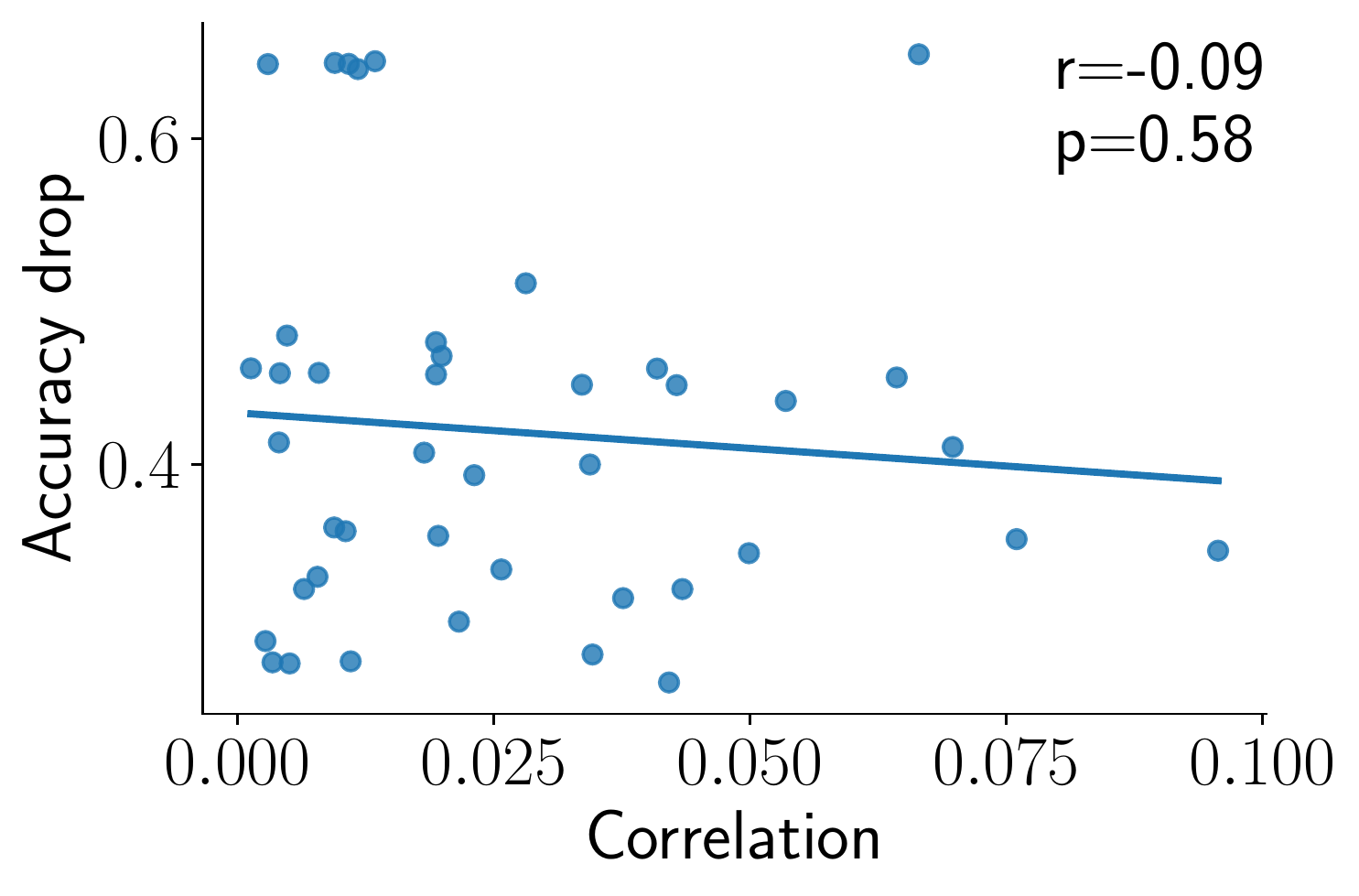} \\
        \bottomrule
    \end{tabular}
    \label{tbl:cor_rel}
\end{table}

Table \ref{tbl:cor_rel} shows the result of our experiment for the Cramér's V and uncertainty coefficient metrics in both opt-out and opt-in. Each point of the figures corresponds to the combination of two attributes (e.g., age and glasses). The x-axis shows the correlation between the two attributes and the y-axis shows the drop in accuracy. We also show the Pearson's correlation coefficient $r$, to measure the nature of the correlation. A value of $0$ would indicate perfectly uncorrelated values, while $-1$ and $1$ indicate perfect negative or positive correlations. Additionally, we calculate the two-tailed $p$-value, which is defined as the probability of an uncorrelated dataset producing this Pearson correlation coefficient. A p-value lower than $0.05$ is generally accepted as statistically significant. These results show that we indeed observe the expected relation between correlation and accuracy drop. Although, the results are not statistically significant enough to make any strong conclusions. 

\subsection{Residual code}
\label{sec:res}
\begin{figure}[h]
\centering
\begin{subfigure}{.5\textwidth}
  \centering
    \includegraphics[width=1.0\textwidth]{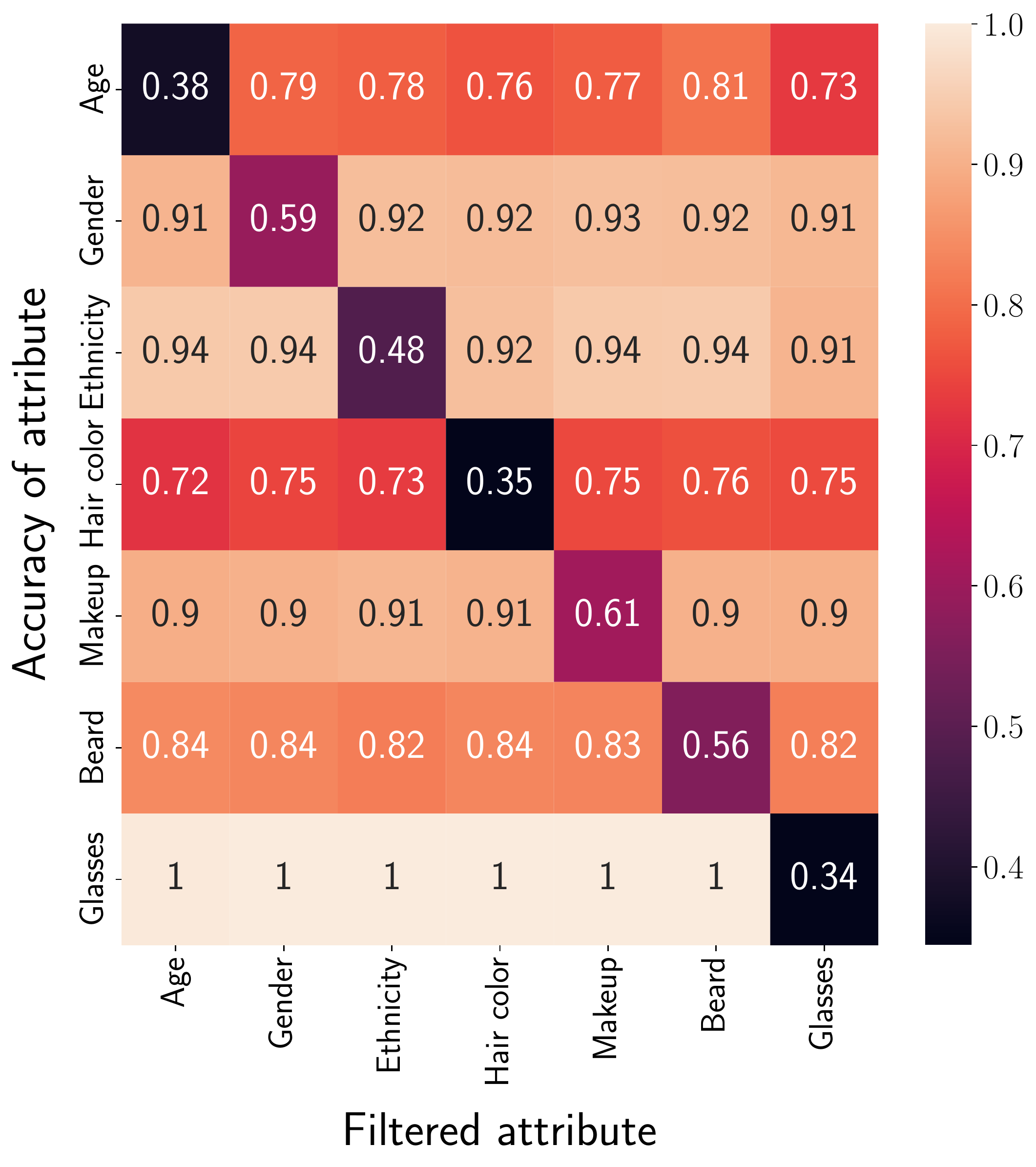}
    \caption{Opt-out}
    \label{fig:acc-res-oo}
\end{subfigure}%
\begin{subfigure}{.5\textwidth}
  \centering
    \includegraphics[width=1.0\textwidth]{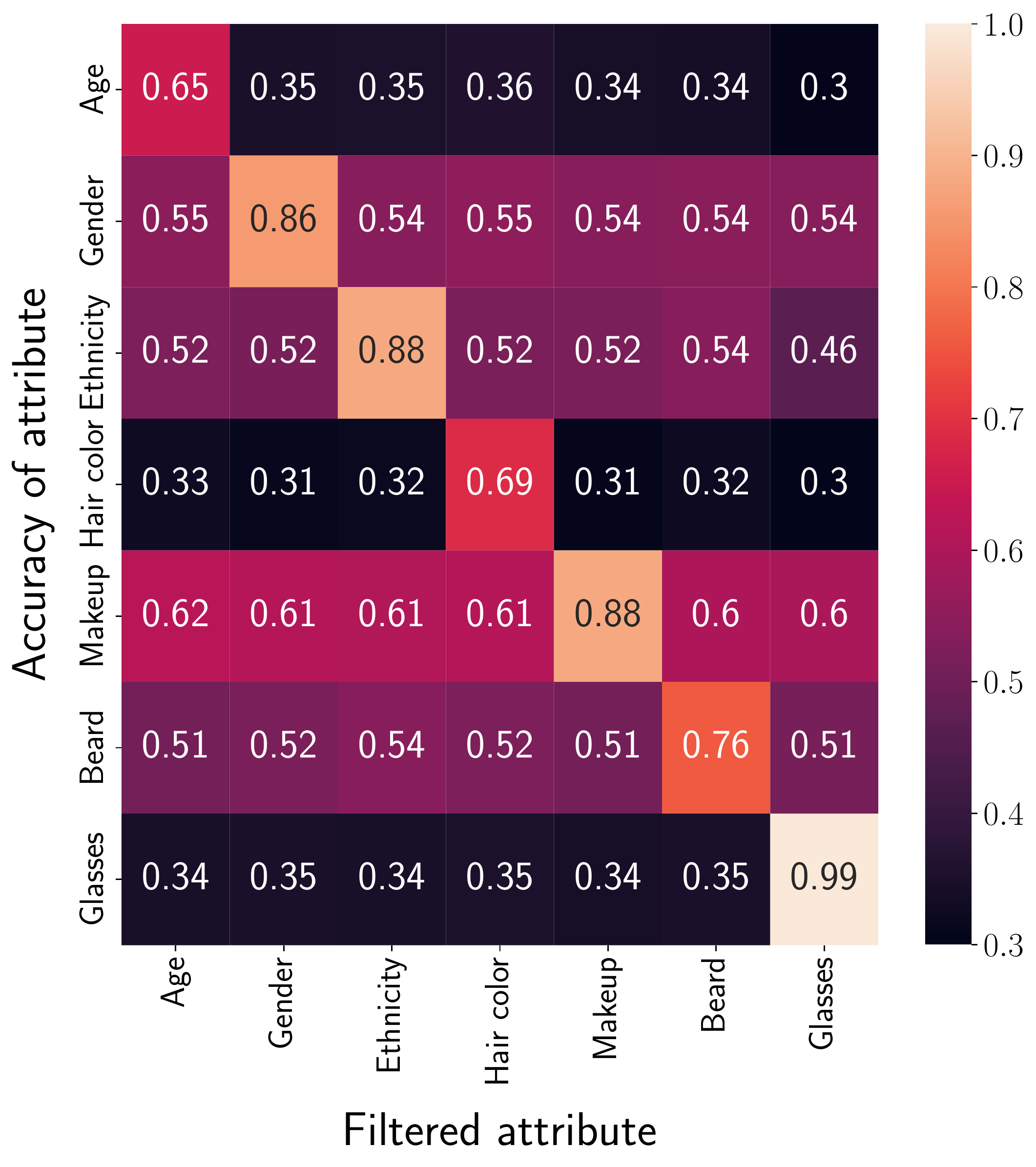}
    \caption{Opt-in}
    \label{fig:acc-res-oi}
\end{subfigure}
\caption{Relative accuracy of filtering when using an opt-out or opt-in configuration on a specified attribute with filtering of the residual codes. Rows indicate the attribute that was filtered, and columns the attribute that was classified.}
\label{fig:acc_residual}
\end{figure}

In Section \ref{sec:filtering} we noticed that the accuracy of filtered attributes remained better than random. A likely cause is that some attribute information is still entangled in the residual code. To determine whether this holds true we repeat the experiment of Section \ref{sec:filtering} with filtered residuals. We chose to filter residual codes by swapping them with those of other images in the dataset to ensure good generation quality by keeping the code in the correct domain. This is in essence true opt-in filtering as no other information than the factor is passed through unchanged. The results of this experiment are visible in Figure \ref{fig:acc_residual}.

Figure \ref{fig:acc_residual} indicates that the residual code affects filtering. Unsurprisingly, removal results in large reductions of accuracy for both the allowed and disallowed attributes. The accuracy of disallowed attributes is now closer to random accuracy. However, those of allowed attributes deteriorated significantly as well, especially when working in an opt-in configuration. These results indicate a clear privacy-utility trade-off.

\section{Generalisation of filtering to unseen attributes}
\label{sec:non_factorized}
\begin{figure}[h]
\centering
\begin{subfigure}[b]{.5\textwidth}
  \centering
    \includegraphics[width=1.0\textwidth]{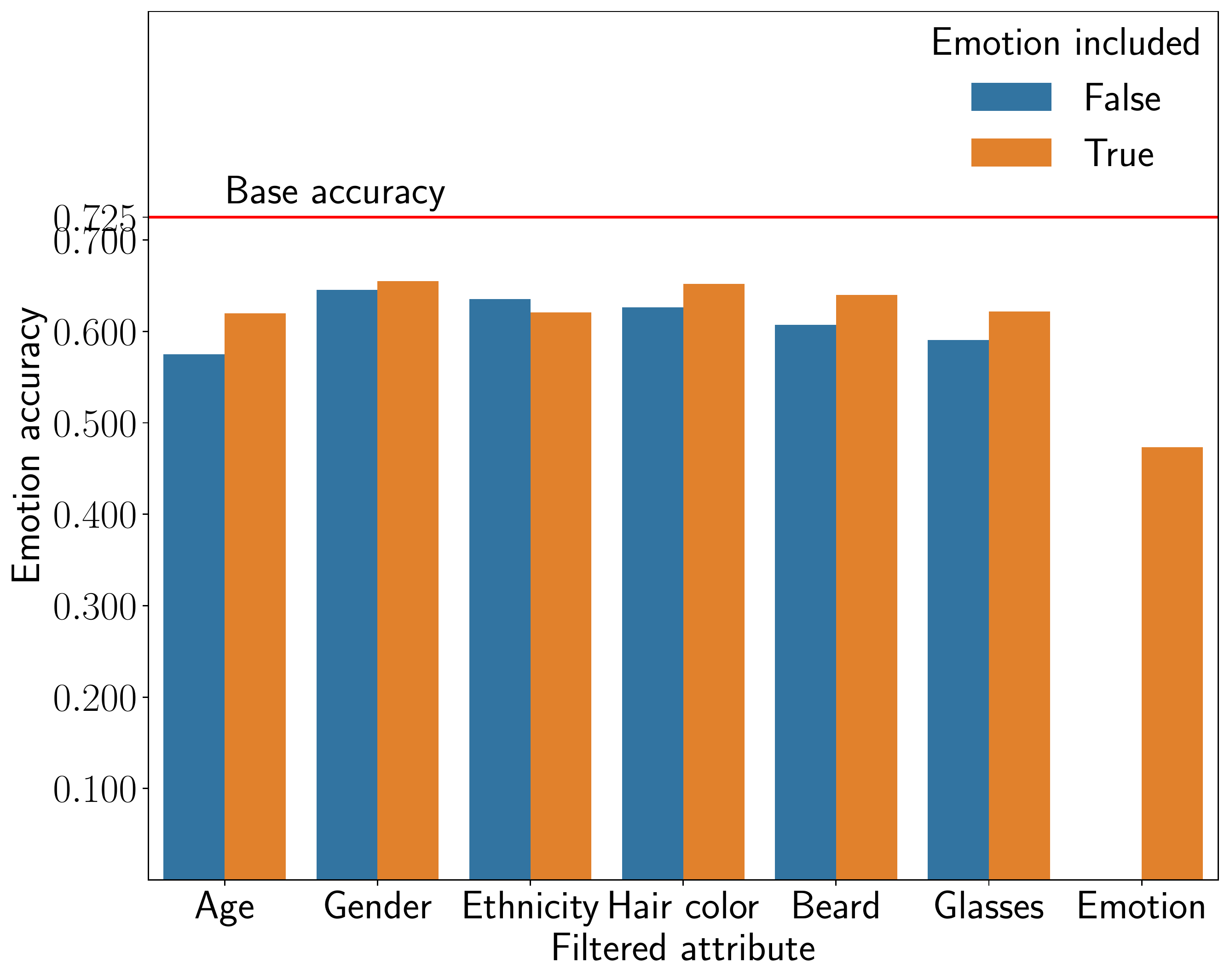}
    \caption{Opt-out}
    \label{fig:emotion-oo}
\end{subfigure}%
\begin{subfigure}[b]{.5\textwidth}
  \centering
    \includegraphics[width=1.0\textwidth]{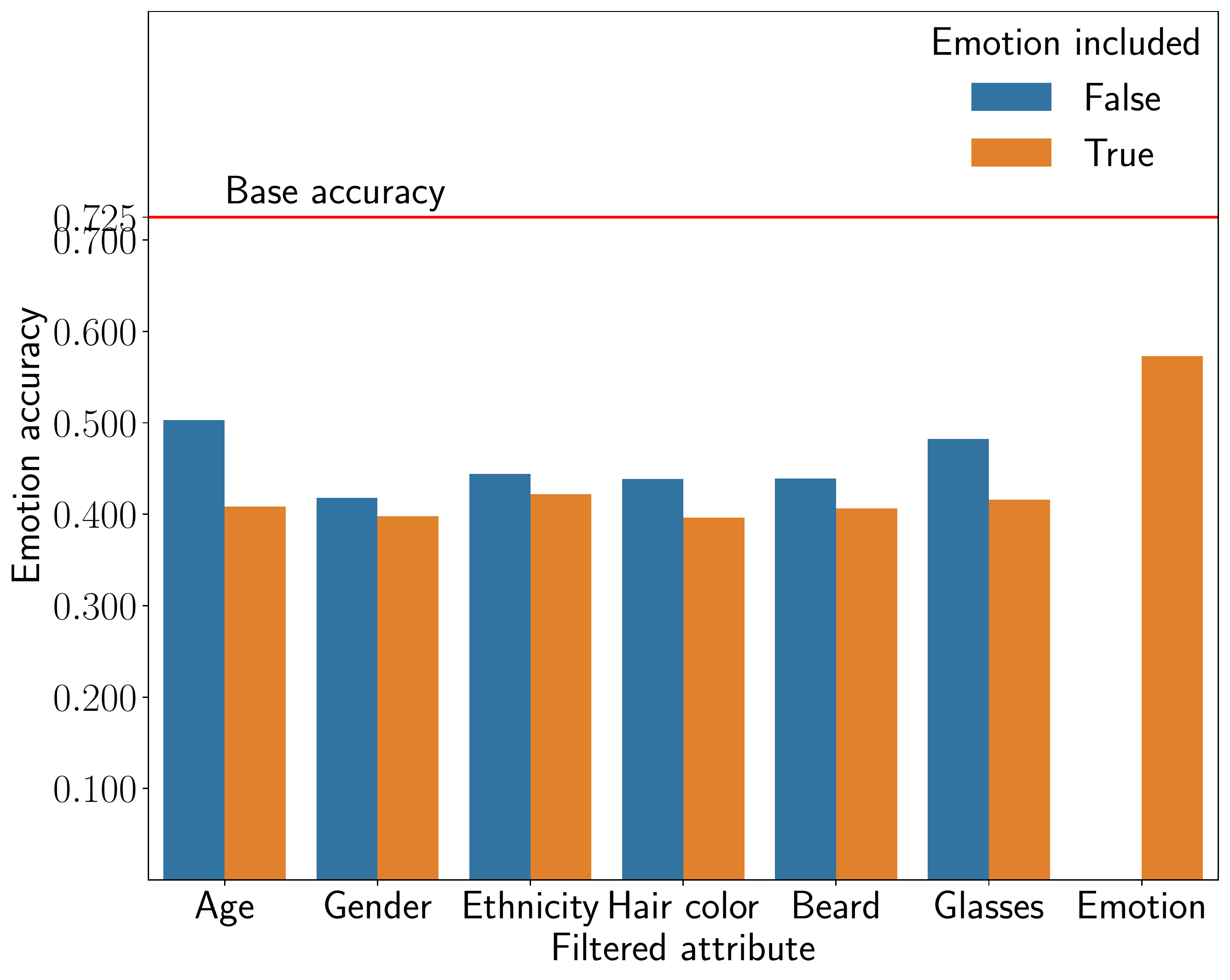}
    \caption{Opt-in}
    \label{fig:emotion-oi}
\end{subfigure}
\caption{Emotion accuracy when filtering other attributes using a ZeroDIM model that included emotion as an attribute and one that did not.}
\label{fig:emotions}
\end{figure}

As a final study, we report on the effects of filtering on non-disentangled attributes. Say, for example, you want to filter out a person's age in an opt-out manner. In principle, this should still allow you to infer that person's gender or hair color,  but it also should pass through facial expressions and other non-age attributes that the filter was not explicitly trained to remove/retain. To highlight the effects of the inclusion of an attribute to filtering, we trained an additional ZeroDIM model, where emotion is included as one of the disentangled attributes. We then evaluated emotion recognition accuracy for every filtering configuration and compared the model's results with and without emotion. We use DAN~\cite{wenDistractYourAttention2021} as our emotion recognition model, which was pre-trained on the AffectNet~\cite{mollahosseiniAffectNetDatabaseFacial2019} dataset. For evaluation, we use the RAVDESS dataset~\cite{livingstoneRyersonAudioVisualDatabase2018}. On  unaltered frames, DAN can achieve an accuracy of $72.45\%$. The results on filtered frames can be seen in Figure \ref{fig:emotions}.

The opt-out results indicate only a slight difference between the model with and without emotion. This leads us to believe that emotion is captured well in the residual code when not specified as a disentangled attribute. In the opt-in setting, we see clearly that we can retain the emotion better when emotion is included as a classifier. When removing emotion, results are again very similar between the two models. However, if an opt-in configuration is needed the model with the attribute included performs significantly better.

\section{Discussion and conclusion}
\label{sec:conclusion}
This paper proposed the use of the state-of-the-art image manipulation model ZeroDIM for filtering face images and studied the implications of natural correlations and residual information on filtering. The findings of this study suggest that image editing is a promising option for both opt-in and opt-out configurations. While the results were positive, we noticed accuracy of filtered attributes was not yet perfectly random. We hypothesized that this was due to two main causes. 

The first is that some attributes are naturally correlated, making it so you cannot filter out one while keeping the other. While we were able to verify correlations between the facial attributes, the results of our experiment were not concise enough to draw any conclusions as to what effects these have on filtering. Further research is thus necessary to confirm or deny our hypothesis. 

The second cause we identified is in the use of the residual code. We confirmed that removing this code led to better filtering. However, it also resulted in lower classification accuracy for all other attributes indicating a privacy-utility trade-off. We could attribute this to the strength of disentanglement as some attribute information clearly remains entangled in the residual code. Though another possibility is that the generator has some implicit dependencies between residual codes and factor codes making modification of the residual incompatible. A possible solution for this would be to create a general residual that allows good generation when swapping codes. This problem of excess information in the residual code becomes highlighted as we found it can capture non-disentangled elements such as emotion quite well.

\section{Future work}
\label{sec:future}
Future work will focus on tackling the investigated limitations of filtering: correlations between the attributes, and residual information. Firstly, we will further investigate the limits correlated attributes put on filtering and how this information can be incorporated to improve filtering. Secondly, we aim to improve the disentanglement of the residual code from the predefined attributes. By doing so, we can decrease information leakage. Furthermore, we aim to investigate the use of a general residual code for filtering. 

To better integrate our application in real-world scenarios, we would further focus on three key aspects: compatibility with third-party API's, eligibility for edge-based cameras, and extension to video data. We would verify if the filtered images can be compatible with existing API's and models such as the Microsoft Face API and FairFace~\cite{karkkainenFairfaceFaceAttribute2021} model. Suitability for resource-constrained devices would be ensured by reducing the computational cost of our filter and investigating its usage in trusted execution environments. Lastly, we plan to expand our technique to video footage and analyze the implications on privacy when a filter operation is performed over multiple frames. 

\section*{Acknowledgments}

This research received funding from the Flemish Government under the ``Onderzoeksprogramma Artifici\"ele Intelligentie (AI) Vlaanderen'' programme.
%
%
%
\bibliographystyle{splncs04}
\bibliography{main}
\end{document}